\documentclass[10pt,twocolumn,letterpaper]{article}

\usepackage{iccv}
\usepackage{times}
\usepackage{epsfig}
\usepackage{graphicx}
\usepackage{amsmath}
\usepackage{amssymb}
\usepackage{algorithm}
\usepackage{algorithmic}
\usepackage{mathrsfs}
\usepackage{multirow}

\usepackage[]{animate}

\usepackage{graphicx}
\usepackage{animate}

\usepackage{float}

\usepackage{multirow}
\usepackage{rotating}
\usepackage{wrapfig}
\usepackage{lineno}
\usepackage{epsfig}
\usepackage{amsmath,amssymb}
\usepackage{color}
\usepackage{booktabs}       
\usepackage{amsfonts,amsmath}       
\usepackage{nicefrac}       
\usepackage{microtype} 
\usepackage{float}
\usepackage{amsmath}

\usepackage{enumerate}
\usepackage{enumitem}

\usepackage[pagebackref=true,breaklinks=true,letterpaper=true,colorlinks,bookmarks=false]{hyperref}

\iccvfinalcopy 

\def\iccvPaperID{2957} 

\ificcvfinal\fi
\begin{document}

\title{Recursively Conditional Gaussian for Ordinal Unsupervised Domain Adaptation}

\author{Xiaofeng Liu$^{1\dag*}$, Site Li$^{2\dag}$, Yubin Ge$^{3}$, Pengyi Ye$^{1}$, Jane You$^{4}$, Jun Lu$^{5}$ \\~\\

$^{1}$Harvard University, Cambridge, MA, USA. \\
$^{2}$Carnegie Mellon University, Pittsburgh, PA, USA. \\
$^{3}$Dept. of Computer Science, University of Illinois at Urbana-Champaign, Urbana, IL, USA.\\
$^{4}$Dept. of Computing, The Hong Kong Polytechnic University, Hong Kong.\\
$^{5}$Beth Israel Deaconess Medical Center, Harvard Medical School, Boston, MA, USA.\\
{\tt\small $^{\dag}$Contribute Equally,  $^{*}$Corresponding Author: liuxiaofengcmu@gmail.com}
}

\maketitle

\begin{abstract}

The unsupervised domain adaptation (UDA) has been widely adopted to alleviate the data scalability issue, while the existing works usually focus on classifying independently discrete labels. However, in many tasks (e.g., medical diagnosis), the labels are discrete and successively distributed. The UDA for ordinal classification requires inducing non-trivial ordinal distribution prior to the latent space. Target for this, the partially ordered set (poset)  is defined for constraining the latent vector. Instead of the typically i.i.d. Gaussian latent prior, in this work, a recursively conditional Gaussian (RCG) set is adapted for ordered constraint modeling, which admits a tractable joint distribution prior. Furthermore, we are able to control the density of content vector that violates the poset constraints by a simple ``three-sigma rule". We explicitly disentangle the cross-domain images into a shared ordinal prior induced ordinal content space and two separate source/target ordinal-unrelated spaces, and the self-training is worked on the shared space exclusively for ordinal-aware domain alignment. Extensive experiments on UDA medical diagnoses and facial age estimation demonstrate its effectiveness.
 

\end{abstract}


\section{Introduction}

Deep learning is typically data-starved and relies on the $i.i.d$ assumption of training and testing sets \cite{che2021deep,liu2021Generalization}. However, the real-world deployment of target tasks is usually significantly diverse, and massive labeling of the target domain data can be expensive or even prohibitive \cite{liu2021adversarial}. To address this, the unsupervised domain adaptation (UDA) is developed, which proposes to learn from both labeled source domain and unlabeled target domain \cite{ganin2016domain,liu2021adapting,liu2021generative}.

\begin{figure}
  \centering
\includegraphics[width=8cm]{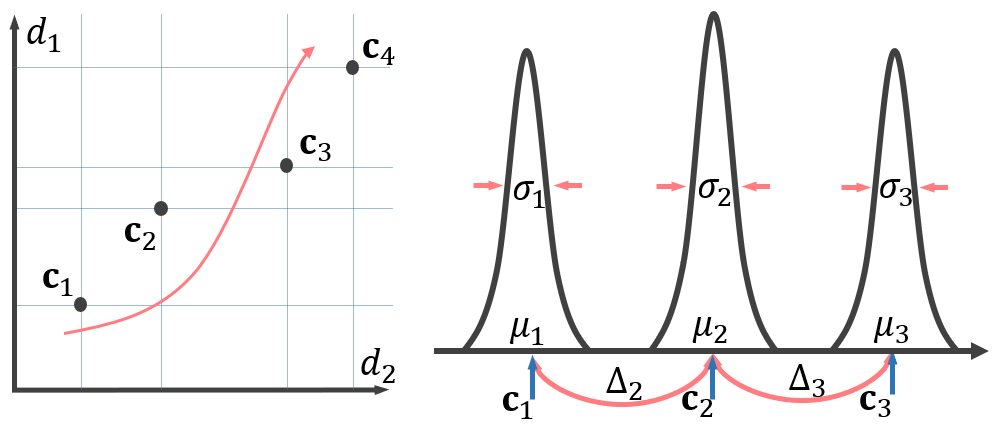}
  \caption{(\textbf{Left}) Illustration of 4-class latent vectors (points) aligned in a poset formation in $\mathbb{R}^2$. For a vector ${\bf c}_{k}$, its adjoined horizontal and vertical lines specify the feasible quadrant where its superiors ${\bf c}_{k+i}$ for $i>0$ can be positioned. 
(\textbf{Right}) Conditional spacing model $p({\bf c}_1,{\bf c}_2,{\bf c}_3) = p({\bf c}_1) p({\bf c}_2|{\bf c}_1) p({\bf c}_3|{\bf c}_2)$ for 3-class. The adopted Gaussian conditional densities (\ref{eq:pv_olvae_conds_1}--\ref{eq:pv_olvae_conds_4}) and the constraints \eqref{eq:pv_param_constrs} enforces the sample from $p({\bf c}_1,{\bf c}_2,\dots,{\bf c}_K)$ satisfy ${\bf c}_1 \leq {\bf c}_2 \leq \dots \leq {\bf c}_K$ with high probability \cite{macneille1937partially,fattore2016partially,kim2020ordinal}.}\label{fig1}  
\end{figure}

The existing UDA works \cite{liu2021subtype,zou2019confidence,liu2021energy,he2020classification,liu2021adapting} usually focus on the classification or segmentation tasks without considering the inter-class correlations. However, tasks with discrete and successively labels are commonly seen in the real world. For instance, the Diabetic Retinopathy (DR) has five labels corresponding to different severity levels: 0$\rightarrow$normal, 1$\rightarrow$mild, 2$\rightarrow$moderate, 3$\rightarrow$severe, and 4$\rightarrow$proliferative DR. In ordinal setting, class $y=1$ is closer to $y=2$ than $y=4$. The misclassification of $y=1$ to $y=2$ or $y=4$ may lead to different severity of misdiagnosis. The ordinal labeling system is widely adopted in medical diagnose, face age groups, and facial expression intensity, etc. Limited by the collecting cost, the UDA has great potential in the medical area, while the UDA for the ordinal label is under-presented.

A promising solution for UDA is to jointly learn the disentanglement of class-related/unrelated factors and the adaptation on the class-relevant latent space exclusively \cite{zou2020joint}. The well-performed model is expected to extract the class label informative and domain invariant features \cite{zou2019confidence}. However, the disentanglement and adaptation of ordinal factors can be challenging \cite{mlvae}. The $i.i.d.$ Gaussian distribution sets are usually adopted as the prior of both disentangled class-related and unrelated latent space \cite{mlvae}, which inherently postulates that the class is independently categorical.

In contrast to the $i.i.d.$ categorical assumption in the conventional disentanglement framework MLVAE \cite{mlvae} with the vanilla Gaussian prior in VAEs \cite{kingma2013auto}, we propose a different setting that the extracted class-related factor ${\bf c}$ is also ordinal-valued w.r.t. the corresponding ordinal label $y$. In other words, the value of class-related factor ${\bf c}$ are also ordered.

We note that if $y$ is an ordinal class label, i.e., $y\in\{1<2<\cdots K\}$, we would expect embed this ordinal structure to form the ordinal class-related latent space. Specifically, for two instances whose class labels are close to each other, their respective extracted ordinal class-related latent vectors ${\bf c}_1$ and ${\bf c}_2$ are expected to be approximate, and vice versa \cite{kim2020ordinal}. For the triplet $y^1<y^2<y^3$, we have: \begin{align}
||{\bf c}_1-{\bf c}_3||> \text{max}\{||{\bf c}_1-{\bf c}_2||,||{\bf c}_2-{\bf c}_3||\},\label{eq:content-distance-ineq} 
\end{align} which aligns the class-related latent vectors and the ordinal class labels, thus yielding a model following the true data causal inference. 

However, it is non-trivial to induce the ordinal inter-class correlation to the disentangled class-related latent space. Actually, directly enforcing the triplet-constraints in Eq.~\ref{eq:content-distance-ineq} as regularization-term for the variational lower bound of disentanglement framework is almost intractable. For the $D$-dimensional ${\bf c}$, there is $\mathcal{O}(D^2)$ inequalities for every pair of $({\bf c}_i,{\bf c}_j)$, and for each inequality, we may have to introduce a slack variable to be optimized \cite{kim2020ordinal}. Moreover, this heuristic regularization cannot guarantee Eq.~\ref{eq:content-distance-ineq} is satisfied in the embedded latent space.

Motivated by the aforementioned insights, we adopted an effective and principled manner for ordinal class-related factor modeling by constructing an appropriate prior for the extracted class-related latent space. Rather than the $i.i.d.$ standard Gaussian prior \cite{kingma2013auto} to the non-ordinal disentanglement methods \cite{mlvae}, our prior is constructed by a recursively conditional Gaussian (RCG) set, which explicitly imposes a restrictive partially ordered set (poset) \cite{macneille1937partially,fattore2016partially,kim2020ordinal} constraint. Since the joint prior is Gaussian, although with a full covariance, inheriting the closed form of the KL divergence term in the variational objective. Moreover, the prior can be fully factorized over the latent dimensions, which make the framework computational tractable. Furthermore, the number of parameters in the adapted prior needed to enforce the constraint for all of the pairs in a $K$ class task can be $\mathcal{O}(D\cdot K)$, further reinforcing the model’s scalability \cite{kim2020ordinal}. This prior is able to assign negligible density that violates the poset constraint by a simple ``three-sigma rule". For the noisy labeled dataset, the confidence interval can be flexibly adjusted to achieve more conservative ordinal modeling.

Our contributions are summarized as follows: 
   
$\bullet$ We propose to investigate the UDA of ordinal classification in a joint disentanglement and adaptation framework.  
   
$\bullet$ An effective and principled ordinal prior for class-related latent space is constructed by a recursively conditional Gaussian (RCG) set \cite{macneille1937partially,fattore2016partially,kim2020ordinal}. More appealingly, we can adjust the ratio of poset violation to adapt the noise of the ordinal label. The closed form of the KL divergence with our adapted prior is computationally tractable for large-scale tasks.

$\bullet$ We extensively evaluate our method on both medical diagnosis (i.e., DR and CHD) and age estimation, and observe a significant improvement over the previous non-ordinal approaches in UDA.

\section{Related works} 

\noindent \textbf{Ordinal classification/regression} can be processed as either multi-class classification task \cite{geng2007automatic} or metric regression task \cite{fu2008human}. The former regards the categorizes as irrelevant to each other, which ignores the inherent ordering correlation among the classes \cite{beckham2017unimodal,liu2018constrained}. The latter assumes the neighboring classes are equally distant, which can violate the non-stationary properties and easily leading to over-fitting \cite{chang2011ordinal}. The threshold-based method proposes to define $K-1$ binary decision boundaries \cite{liu2018ordinal}. Recently, the Wasserstein loss is adopted as an alternative to conventional cross-entropy loss to incorporate the inter-class correlations \cite{liu2019unimodal,liu2020wasserstein,han2020wasserstein,liu2020severity,liu2020importance,liu2019conservative,ge2021embedding}. The restrictive partially ordered set (poset) has also been used to enforce the ordinal latent representation \cite{fattore2016partially,kim2020ordinal}. The application of UDA for ordinal data is even more promising than the conventional task since the ordinal data is usually hard to label \cite{liu2018ordinal}. Our method is also orthogonal to the recent ordinal classification loss \cite{liu2019unimodal}.

\noindent \textbf{Unsupervised domain adaptation} is aimed at improving the performance in an unlabeled target domain with the help of a labeled source domain \cite{kouw2018introduction}. The typical methods can be mainly grouped into statistic moment matching \cite{long2018conditional}, domain style transfer \cite{sankaranarayanan2018generate,he2020image2audio}, feature-level adversarial learning \cite{ganin2016domain,tzeng2017adversarial,liu2018data} and self-training \cite{zou2019confidence,liu2021energy,liu2021generative}. A well-performed solution is disentangling the input to class-related/unrelated factors and only aligns the class-related factor across domains \cite{zou2020joint,bousmalis2016domain}. Moreover, the adaptation is achieved by self-training \cite{zou2019confidence}, which infers the pseudo-label for target samples in each iteration to gradually reduce the cross-domain variations. 

However, it can be challenging to induce the ordinal prior in the disentangled latent space \cite{liu2019unimodal,liu2020unimodal,fattore2016partially,kim2020ordinal}. We propose a joint disentangling and adaptation framework based on VAEs and tailors the prior for ordinal class-related factor (RCG) and class-unrelated factors (Gaussian).  

We also tried the combination of Wasserstein loss \cite{liu2019unimodal} with self-training \cite{zou2019confidence,zou2020joint}, while the ordinal property of the to be transferred feature can not be explicitly enforced. Moreover, it is commonly challenging to combine loss-related UDA methods \cite{pan2019transferrable} with the ordinal loss \cite{liu2019unimodal}.



\noindent \textbf{Disentangled representation}. Learning the representations that are informative to the label and robust to the other annoying variations is a primary target of representation learning \cite{liu2021mutualpami,liu2019feature,liu2021mutualpr,liu2017deep}. Although some works \cite{infogan16,beta_vae17,tcvae} shown that unsupervised disentanglement can be achieved by augmenting the loss of VAEs~\cite{vae14} also to enforce the marginal independence of the prior imposed by the encoders. While it is shown that besides sufficient inductive bias or regularity constraint is induced, the unsupervised disentanglement suffers from unidentifiability~\cite{impossibility}. Actually, in a self-training-based UDA setting, we have the ground truth class label of source domain data, and the pseudo-label of target domain data is inferred. The group-level MLVAE model~\cite{mlvae} is an extension of VAEs~\cite{vae14}, which assumes a sample ${\bf x}$ from class $y=k$ is a generation result of a pair of latent vectors ${\bf c}_k$ and ${\bf u}$ for content and instance-specific style, respectively. Therefore, all of the samples ${\bf x}$ from class $y=k$ should have the same ${\bf c}_k$ for content, while ${\bf u}$ summarizes the other complementary variations of each sample. In contrast to MLVAE~\cite{mlvae}, we target the {\em ordinal-valued} ${\bf c}_k$, in which the values of content factors are {\em ordered} rather than categorical. Moreover, the ordinal prior induced VAEs is further incorporated in an adaptation framework. 

Several recent VAE studies attempt to process the ordinal labeled data, while their objectives and setups, on the other hand, fundamentally differ from our framework. \cite{ordinal_vae} assumes the ordinal paired samples are accessible for training, which is different from the ordinal classification task \cite{liu2018ordinal}. The variational posterior of the ordinal class is introduced in \cite{ordinal_vae2}. \cite{granger2020deep} utilizes the video-level label in the target domain and does not require the extracted latent vectors are able to align the ordinal constraint.

\section{Methodology}

In ordinal UDA, given a source domain $p({\bf x}_s,y)$ and a target domain $p({\bf x}_t,y)$, a labeled set $\mathcal{D_{S}}$ is drawn $i.i.d$ from $p({\bf x}_s,y)$, and an unlabeled set $\mathcal{D_{T}}$ is drawn $i.i.d$ from the marginal distribution $p({\bf x}_t)$. The goal of UDA is to build a good classifier in the target domain with the training on $\mathcal{D_{S}}$ and $\mathcal{D_{T}}$. $y\in\left\{1,2,\dots,K\right\}$ is the discrete and ordinal class label. In self-training UDA \cite{zou2019confidence}, we infer the pseudo-label $\hat{y}$ for target samples in each iteration. Given a group of source and target images, we assume there is a ordinal prior induced shared space for the class-related factors and two domain-specific class-unrelated spaces, and the self-training \cite{zou2019confidence} is applied on the shared ordinal class-related space.

For a group of source images $\{{\bf x}_s^n\}_{n=1}^{N_s}$ with $y=k$ and target images $\{x_t^n\}_{n=1}^{N_t}$ with $\hat{y}=k$, where $n$ indicates a group of cross-domain data instance, $N_s$ and $N_t$ denote the source and target domain sample numbers. Our proposed framework, is built with the VAEs~\cite{kingma2013auto,mlvae} backbones to encode a sample ${\bf x}^n_s$ or ${\bf x}^n_t$ to the latent representation pair of $({\bf c}^n,{\bf u}^n_s)$ or $({\bf c}^n,{\bf u}^n_t)$, where ${\bf c}^n$ is responsible for {\em ordinal class-related} and all of the other variations are considered as {\em class-unrelated feature} ${\bf u}^n_s$ or ${\bf u}^n_t$ for source or target domains respectively. The change of ${\bf c}^n$ is expected to exclusively affect the ordinal class-related aspect and, conversely, the 
class-unrelated ${\bf u}^n_s$ or ${\bf u}^n_t$ is independent of ${\bf c}^n$. We note that the class-unrelated factors are not labeled \cite{liu2019feature}.


\subsection{Cross-domain Ordinal Graphical Model}\label{sec:principled} 

By the definition of class-related ${\bf c}^n$, we expect the ${\bf c}^n$ extracted from the samples within the same class should be similar or identical. Therefore, it is reasonable to calculate a representative feature ${\bf c}_k$ based on a group of ${\bf c}^n$ from the same class $k$ as in \cite{mlvae}. Practically, we can use the average or the simple product-of-expert rule adopted in MLVAE \cite{mlvae}. 

For the well disentangled and adapted features, the observed data instance in source domain ${\bf x}^n_s$ with $y^n=k$ (or target domain ${\bf x}^n_t$ with $\hat{y}^n=k$) can be regarded as a generation result based on a pair of latent vectors $({\bf c}_k, {\bf u}^n_s)$ (or $({\bf c}_k, {\bf u}^n_t)$), respectively. We note that the ordinal label $y$ or the pseudo label $\hat{y}$ predicted by the self-training is always available in each iteration. This differentiates ${\bf c}^n$ from ${\bf u}_s^n$, ${\bf u}_t^n$ in that there is one ${\bf c}_k$ that governs all cross-domain instances with the same ordinal class $y=k$. However, each instance ${\bf x}^n_s$ (or ${\bf x}_t^n$) have different random variable ${\bf u}^n_s$ (or ${\bf u}_t^n$), which usually contains the source/target domain exclusive information and is more reasonable to configure two domain-specific VAEs. 

We provide the graphical model with plate notation in Fig.~\textcolor{red}{2}. $p({\bf x}_s^n | {\bf c}_k, {\bf u}_s^n)$ and $p({\bf x}_t^n | {\bf c}_k, {\bf u}_t^n)$ are the source and target decoders, which generate images from the class-wise ordinal-related and domain-wise ordinal-unrelated latent.

 \begin{figure}[t]
  \centering
\includegraphics[height=3.8cm]{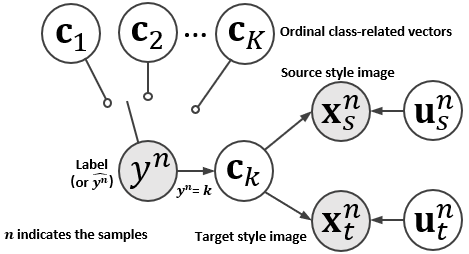} 
  \caption{Graphical model representation of the proposed framework for ordinal UDA. The gray-shaded circles indicate the observed variables of the sample ${\bf x}$ and its ordinal label ${y}$. }\label{fig2}  
\end{figure}

\subsection{Latent Priors}\label{sec:latent_priors} 

We configure three encoders $Enc^c_{s,t}=q^c_{s,t}({\bf c}^n|{\bf x}_{s,t}^n)$, $Enc_s^u=q^u_s({\bf u}_s^n|{\bf x}_s^n)$ and $Enc_t^u=q^u_t({\bf u}_t^n|{\bf x}_t^n)$ for latent space inference, and two domain-specific decoders $Dec_s^u=p_s({\bf x}_s^n|{\bf c}_k,{\bf u}_s^n)$, $Dec_t^u=p_t({\bf x}_t^n|{\bf c}_k,{\bf u}_t^n)$ to generate the image with sampled latent pairs as shown in Fig.~\textcolor{red}{3}.

For the prior of class-unrelated factors, a natural choice is the $i.i.d.$ standard Gaussian $\mathcal{N}({\bf 0}, {\bf I})$. Specifically, we can set $p(\{{\bf u}_s^n\}_{n=1}^{N_s}) = \prod_{n=1}^{N_s} \mathcal{N}({\bf u}^n_s; {\bf 0}, {\bf I})$ and $p(\{{\bf u}_t^n\}_{n=1}^{N_t}) = \prod_{n=1}^{N_t} \mathcal{N}({\bf u}^n_t; {\bf 0}, {\bf I})$ following vanilla VAEs~\cite{kingma2013auto}. As for the class-related reference vectors, a possible choice is following the MLVAE~\cite{mlvae} to use fully factorized standard Gaussian distributions, $p({\bf c}_1,\dots,{\bf c}_K) = \prod_{k=1}^K \mathcal{N}({\bf c}_k; {\bf 0}, {\bf I})$. It should be noted that MLVAE's $i.i.d.$ prior modeling of the different class is meaningful only if the label values are categorical (i.e., discrete and independent).

Moreover, for the ordinal label, where the approximate ordinal classes are expected to be encoded. However, the $i.i.d.$ Gaussian prior can be suboptimal, since it can not maintain the ordinal feature. Target for a more reasonable ordinal content prior model, we expect the triplet ordering constraint in Eq.~\ref{eq:content-distance-ineq} can be imposed. A possible manner of introducing this constraint is to simply adopt it as a regularization term. However, this heuristic design can not guarantee the inequity in Eq.~\ref{eq:content-distance-ineq}, and not be scalable to a large dataset.

Following \cite{fattore2016partially,kim2020ordinal}, we propose to induce this constraint by having a {\em partially ordered set (poset)} \cite{macneille1937partially} for the ordinal class-related latent. For each dimension $d\in\{1,\cdots,D\}$, we propose to order the $d_{th}$ element in the latent vectors, i.e., $[{\bf c}_1]_d < [{\bf c}_2]_d < \cdots < [{\bf c}_K]_d$, where $[{\bf c}_k]_d$ is the $d$-th entry of the vector ${\bf c}_k$ (i.e., ${\bf c}_k = [ [{\bf c}_k]_1, [{\bf c}_k]_2, \dots, [{\bf c}_k]_d ]^\top$). These vectors align in a poset and satisfy the ordinal class related latent-distance constraints in Eq.~\ref{eq:content-distance-ineq}. The case of $D=2$-dim latent space is shown in Fig.~\textcolor{red}{1} left.

%

\begin{figure}
  \centering
\includegraphics[width=9cm]{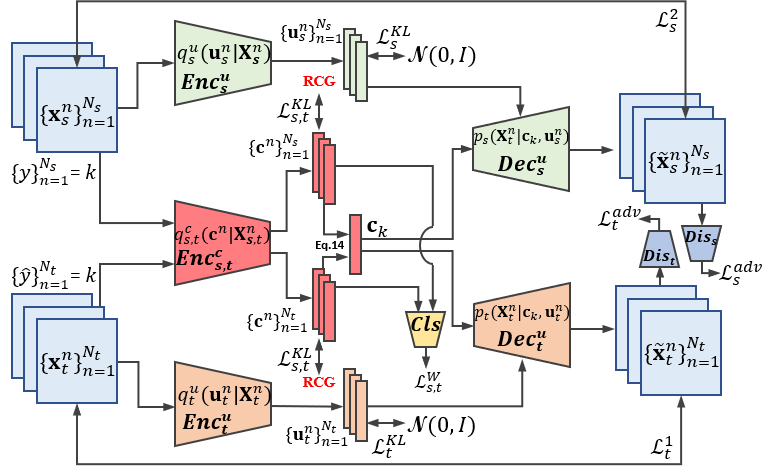} 
  \caption{Our proposed ordinal UDA framework with a Recursively Conditional Gaussian (RCG) induced disentangled ordinal-related latent space. Only $Enc^c_{s,t}$ and $Cls$ are used in testing.}\label{fig3} 
\end{figure}

For imposing the poset constraints in the prior, we resort to the {\em recursively conditional Gaussian} (RCG) set \cite{fattore2016partially,kim2020ordinal}. The framework imposes a joint Gaussian distribution over the $K$ vectors, fully correlated for $k\in\{1,\cdots,K\}$, and fully factorized for dimensions $d\in\{1,\cdots,D\}$, enables the variational inferences to be tractable for computation. It is possible to effectively place negligible RCG densities that not meet the poset constraints. Specifically, we consider a dimension-wise independent distribution, \begin{equation}
p({\bf c}_1,\dots,{\bf c}_K) = \prod_{d=1}^D p_d([{\bf c}_1]_d, \dots, [{\bf c}_K]_d),
\label{eq:pv_olvae_dimwise_indep} 
\end{equation}
where $p_d$ is the density over $K$ variables from $d$-th dimension of the latent vectors. 
We model $p_d$ by a product of predecessor-conditioned Gaussians. 
For simplicity, we drop the subscript $d$ in notation, and abuse $c_k$ to denote $[{\bf c}_k]_d$, $p(c_1,\dots,c_K)$ to refer to $p_d([{\bf c}_1]_d, \dots, [{\bf c}_K]_d)$. Following \cite{macneille1937partially,fattore2016partially,kim2020ordinal}, the framework can be formulated as: 
\begin{equation}
p(c_1,\dots,c_K) = p(c_1) p(c_2|c_1)\cdots p(c_K|c_{K-1}), 
\label{eq:pv_olvae_conditioned} 
\end{equation}
where the conditionals are defined as:
\begin{align}
p(c_1) &= \mathcal{N}(c_1; \mu_1, \sigma_1^2) \label{eq:pv_olvae_conds_1} \\ 
p(c_2|c_1) &= \mathcal{N}(c_2;  \underbrace{c_1+\Delta_2}_{:=\mu_2}, \sigma_2^2) 
\label{eq:pv_olvae_conds_2} \\
p(c_3|c_2) &= \mathcal{N}(c_3;  \underbrace{c_2+\Delta_3}_{:=\mu_3}, \sigma_3^2) 
\label{eq:pv_olvae_conds_3} \\ 
& \ \ \vdots \nonumber \\ 
p(c_K|c_{K-1}) &= \mathcal{N}(c_K;  \underbrace{c_{K-1}+\Delta_K}_{:=\mu_K}, \sigma_K^2).
\label{eq:pv_olvae_conds_4} 
\end{align}
We define $p(c_k|c_{k-1})$ as a Gaussian centered at $\mu_k := c_{k-1} + \Delta_k$ with variance $\sigma_k^2$. That is, $\Delta_k$ ($> 0$) is the spread between the predecessor sample $c_{k-1}$ and the mean $\mu_k$ as shown in Fig.~\textcolor{red}{1} right. 
We consider $\{\sigma_k,\Delta_k\}_{k=1}^K$ and $\mu_1$ to be the free parameters of the framework that can be learned from data.

To efficiently approximate (in other words, roughly to guarantee) the poset constraint, we make each conditional distribution (pillar) separated from its adjacent neighbors through the following ``three-sigma" constraint:  
\begin{equation}
\Delta_k \geq 3\sigma_k, 
\label{eq:pv_param_constrs} 
\end{equation}
which has $>99.7\%$ probability as near certainty \cite{pukelsheim1994three}. With Eq.~\ref{eq:pv_param_constrs}\footnote{
The inequality constraints \eqref{eq:pv_param_constrs} can be easily incorporated in the conventional unconstrained optimizer modules such as PyTorch and TensorFlow through trivial reparametrizations \cite{kim2020ordinal}, e.g., $\sigma_k := \frac{\Delta_k}{3} \textrm{sigmoid}(\overline{\sigma}_k)$ and $\Delta_k := \exp(\overline{\Delta}_k)$, where $\overline{\Delta}_k$ and $\overline{\sigma}_k$ are the unconstrained optimization variables, and $\textrm{sigmoid}(x) = 1/(1+\exp(-x))$. 
Furthermore, we fix $\sigma_1=1$ to make the optimization numerically more stable.
}, the likelihood that $c_k \leq c_{k-1}$ is negligible
enforcing the desired ordering $c_1 < c_2 < \dots < c_K$. We note that the label noise is usually significant in many ordinal dataset, and the unimodal label-smoothing is widely used to achieve conservative prediction \cite{liu2019unimodal}. Instead of sophisticated ordinal noise modeling \cite{liu2020unimodal}, we can simply adapt to ``one/two-sigma" constraint (i.e., $\Delta_k \geq 2\sigma_k$) to tolerate the noise, which also provides more than $68\%$ or $95\%$ confidence interval \cite{pukelsheim1994three}.

The joint density for Eq.~\ref{eq:pv_olvae_conds_1}--\ref{eq:pv_olvae_conds_4} has a closed-form. Basically, given that all of them are Gaussian, yet linear, we have the joint density $p(c_1,c_2,\dots,c_K)$ as Gaussian. Therefore, we can formulate the mean and covariance of the full joint Gaussian model as \cite{macneille1937partially,fattore2016partially,kim2020ordinal}: 
\begin{align}
&\mathbb{E}[c_k] = \mu_1 + \Delta_2 + \cdots + \Delta_k \ \ (\textrm{for} \ \ k \geq 2) 
\label{eq:pv_joint_mean} \\
&\mathbb{\textrm{Cov}}(c_i,c_j) = \sigma_1^2 + \cdots +  \sigma_{\min(i,j)}^2 
\label{eq:pv_joint_cov}
\end{align}
The joint distribution $p(c_1,c_2,c_3)$ For $K=3$ is: 
\begin{equation}
\mathcal{N}\Bigg( 
\underbrace{
  \begin{bmatrix}
    \mu_1 \\
    \mu_1+\Delta_2 \\
    \mu_1+\Delta_2+\Delta_3
  \end{bmatrix}
}_{:={\bf a}},
\underbrace{
  \begin{bmatrix}
    \sigma_1^2 & \sigma_1^2 & \sigma_1^2 \\
    \sigma_1^2 & \sigma_1^2 + \sigma_2^2 & \sigma_1^2 + \sigma_2^2 \\
    \sigma_1^2 & \sigma_1^2 + \sigma_2^2 & \sigma_1^2 + \sigma_2^2 + \sigma_3^2
  \end{bmatrix}
}_{:={\bf C}}
\Bigg),
\label{eq:pv_joint_ex} 
\end{equation}
where ${\bf a}$ and ${\bf C}$ are the mean vector and covariance matrix of the joint Gaussian. 
Plugging this back in our original prior model Eq.~\ref{eq:pv_olvae_dimwise_indep}, 
we have: 
\begin{equation}
p({\bf c}_1,\dots,{\bf c}_K) 
= \prod_{d=1}^D \mathcal{N}([{\bf c}]_d; {\bf a}_d, {\bf C}_d),
\label{eq:pv_olvae} 
\end{equation}
where $[{\bf c}]_d := \big[ [{\bf c}_1]_d, \dots, [{\bf c}_K]_d \big]^\top$ is the $K$-dim vector collecting $d$-th dim elements from ${\bf c}_k$. Also, ${\bf a}_d$ and ${\bf C}_d$, for each $d=1,\dots,D$, are defined by Eq.~\ref{eq:pv_joint_mean}--\ref{eq:pv_joint_ex} with their own free parameters, denoted as: $\big( \mu^{d}_{1}, \{\Delta^{d}_k, \sigma^{d}_k \}_{k=1}^K \big)$ \cite{kim2020ordinal}. The covariance ${\bf C}_d$ is not diagonal. However, the model is factorized over $d=1,\dots,D$, the fact exploited in the next section is able to make the variational inference tractable.

\subsection{Variational Inference}\label{sec:vi}

Given the ordinal data $\{({\bf x}_s^n, y^n)\}_{n=1}^{N_s}$, $\{({\bf x}_t^n, \hat{y}^n)\}_{n=1}^{N_t}$, we approximate the posterior by the following variational density, decomposed into the ordinal class-related and unrelated latents. The ordinal class-related posterior is further factorized over the ordinal class levels $y=1,\dots,K$, 
\begin{align}
q_s\big( \{{\bf c}_k\}_{k=1}^K, \{{\bf u}^n_s\}_{n=1}^{N_s} \big) = \prod_{k=1}^K q^c_{s,t}({\bf c}_k|G_k) \prod_{n=1}^{N_s} q_{s}^u\big( {\bf u}^n_s | {\bf x}_s^n \big),\nonumber\\
q_t\big( \{{\bf c}_k\}_{k=1}^K, \{{\bf u}^n_t\}_{n=1}^{N_t} \big) = \prod_{k=1}^K q^c_{s,t}({\bf c}_k|G_k) \prod_{n=1}^{N_t} q_{t}^u\big( {\bf u}^n_t | {\bf x}_t^n \big),
\label{eq:vi_Q_form} 
\end{align}
where $q^c_{s,t}({\bf c}_k|G_k)$ is short notation of $q^c_{s,t}\big( {\bf c}_k|\{{\bf x}^n_{s,t}\}_{n \in G_k} \big)$ and $G_k = \{\{{\bf x}_s^n\}_{n=1}^{N_s},\{{\bf x}_t^n\}_{n=1}^{N_t}\}$ is the set of source and target training instances with ordinal class label $y=k$ or $\hat{y}=k$. For the encoders, we adopt deep networks that take an input sample and output the means and variances of the Gaussian-distributed latents following the reparameterization trick in vanilla VAEs \cite{kingma2013auto}. However, since $Enc^c_{s,t}$ requires a group of samples $\{{\bf x}^n_{s,t}\}_{n\in G_k}$ as its input, instead of adopting a complex group encoder such as the neural statisticians~\cite{nstatistician,homoencoder}, we use a simple product-of-expert rule adopted in MLVAE \cite{mlvae,kim2020ordinal}:  

\begin{algorithm}[t]
\caption{Adaption process of our RCG induced UDA.}
\label{alg:A}
\begin{algorithmic}
\STATE {Initialize parameters of encoders, decoders and classifier.} 
\REPEAT 
\STATE Sample $G_k=\{\{{\bf x}_s^n\}_{n=1}^{N_s},\{{\bf x}_t^n\}_{n=1}^{N_t}\}$ with $y$/$\hat{y}=k$.
\STATE Compute $\{{\bf c}^n\}_{n=1}^{Ns},\{{\bf c}^n\}_{n=1}^{Nt},\{{\bf u}_k^n\}_{n=1}^{Ns},\{{\bf u}_k^n\}_{n=1}^{Nt}$.
\STATE Classify $\{{\bf c}^n\}_{n=1}^{Ns},\{{\bf c}^n\}_{n=1}^{Nt}$ with $Cls$.
\STATE Calculate the loss functions: $\mathcal{L}^{KL}_{s,t}$, $\mathcal{L}^{KL}_s$, $\mathcal{L}^{KL}_t$, $\mathcal{L}^{W}_{s,t}$.
\STATE Construct ${\bf c}_k$ with $\{{\bf c}^n\}_{n=1}^{Ns},\{{\bf c}^n\}_{n=1}^{Nt}$ using Eq. 15.
\STATE Decode $\{\tilde{\bf x}_s^n\}_{n=1}^{Ns}$,$\{\tilde{\bf x}_t^n\}_{n=1}^{Nt}$ and calculate $\mathcal{L}^1_s$, $\mathcal{L}^1_t$.
\vspace{+5pt} 
\STATE //Update parameters according to gradients:
\STATE $Enc^c_{s,t}\leftarrow\mathcal{L}^{W}_{s,t}+\alpha\mathcal{L}^{KL}_{s,t}+\beta\mathcal{L}^1_s+\beta\mathcal{L}^1_t +\gamma\mathcal{L}^{adv}_s +\gamma\mathcal{L}^{adv}_t$;\vspace{-5pt} 
\STATE $Enc^u_s\leftarrow\mathcal{L}^1_s+\lambda\mathcal{L}^{KL}_s+\theta\mathcal{L}^{adv}_s;Dec^u_s\leftarrow\mathcal{L}^1_s+\theta\mathcal{L}^{adv}_s$; \vspace{-5pt} 
\STATE $Enc^u_t\leftarrow\mathcal{L}^1_t+\lambda\mathcal{L}^{KL}_t+\theta\mathcal{L}^{adv}_t;Dec^u_t\leftarrow\mathcal{L}^1_t+\theta\mathcal{L}^{adv}_t$; \vspace{-5pt} 
\STATE $Cls\leftarrow\mathcal{L}^{W}_{s,t}$; $Dis_s\leftarrow-\mathcal{L}^{adv}_{s}$; $Dis_t\leftarrow-\mathcal{L}^{adv}_{t}$.
\UNTIL{deadline}
\end{algorithmic}
\end{algorithm} 

\begin{figure*}[t]
\centering
\includegraphics[width=17.5cm]{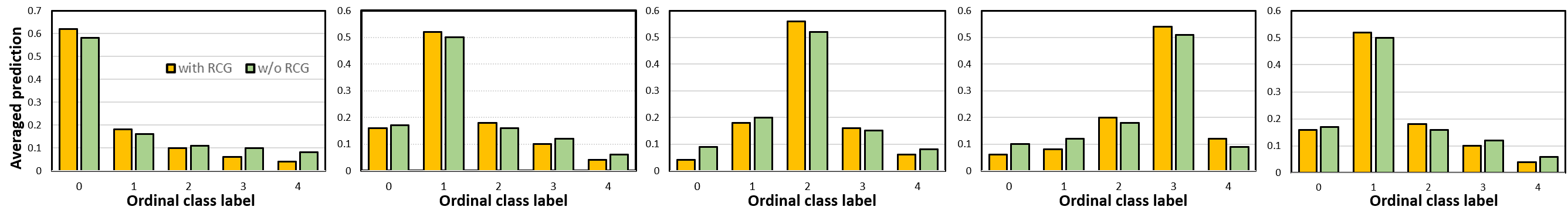}
\caption{The distributions of average predictions on the target testset of DR UDA task. The RCG prior (orange) usually predicts more concentrated output distribution around the ground-truth label than Gaussian prior UDA (green). Best viewed in color.}
\label{fig:example} 
\end{figure*}

\begin{equation}
q^c_{s,t}({\bf c}_k|G_k) \propto \prod_{n \in G_k}  q^c_{s,t}({\bf c}^n | {\bf x}^n_{s,t}).
\label{eq:poe} 
\end{equation}
Since each $q^c_{s,t}({\bf c}^n | {\bf x}^n_{s,t})$ is Gaussian, the product in Eq.~\ref{eq:poe} follows a Gaussian,  and it can be shown that the evidence lower bound (ELBO) can be formulated as:  
\begin{align}
&~~~~\sum_{k=1}^K \mathbb{E}_{q^c_{s,t}({\bf c}_k|G_k)} \sum_{n\in G_k} \mathbb{E}_{q^u_s( {\bf u}^n_s | {\bf x}^n_s)} \Big[ \log p({\bf x}^n_s|{\bf c}_k,{\bf u}^n_s) \Big] \nonumber \\
&+\sum_{k=1}^K \mathbb{E}_{q^c_{s,t}({\bf c}_k|G_k)} \sum_{n\in G_k} \mathbb{E}_{q^u_t( {\bf u}^n_t | {\bf x}^n_t)} \Big[ \log p({\bf x}^n_t|{\bf c}_k,{\bf u}^n_t) \Big] \nonumber \\
&-\sum_{n=1}^{N_s} \textrm{KL}\Big( 
  q^u_s\big( {\bf u}^n_s | {\bf x}^n_s \big) \big\Vert p({\bf u}^n_s) 
\Big)\nonumber\\
&-\sum_{n=1}^{N_t} \textrm{KL}\Big( 
  q^u_t\big( {\bf u}^n_t | {\bf x}^n_t \big) \big\Vert p({\bf u}^n_t) 
\Big)\nonumber\\
&-\textrm{KL}\big( 
  \prod_{k=1}^K q^c_{s,t}({\bf c}_k | G_k) \bigg\Vert p({\bf c}_1,\dots,{\bf c}_K) 
\big). 
\label{eq:elbo} 
\end{align}
where $p({\bf u}^n_s), p({\bf u}^n_t) \sim\mathcal{N}({\bf 0},{\bf I})$ as vanilla VAEs \cite{kingma2013auto}, while $p({\bf c}_1,\dots,{\bf c}_K)$ is the ordinal-constrained prior given in Eq.~\ref{eq:pv_olvae}. Moreover, the first two terms are the cross-domain reconstructions, which is optimized with the L1 loss $\mathcal{L}_s^1=|{\bf x}_s^n-{\tilde{\bf x}}_s^n|$, $\mathcal{L}_t^1=|{\bf x}_t^n-{\tilde{\bf x}}_t^n|$ for all samples, and the two class-unrelated KL divergences $\mathcal{L}_{s}^{KL}$, $\mathcal{L}_{t}^{KL}$ follow the two domain versions of MLVAE \cite{mlvae}, with the key difference in the last term $\mathcal{L}_{s,t}^{KL}$ with induced ordinal prior RCG.

\textbf{Tractable ordinal latent KL-term}. Despite all distributions in the KL term are Gaussians, the full dependency of ${\bf c}_k$'s over $k\in\{1,\cdots,K\}$ in $p({\bf c}_1,\dots,{\bf c}_K)$ can be illposed if both $D$ and $K$ are large \cite{macneille1937partially,fattore2016partially,kim2020ordinal}. For the $D\cdot K$ dimension input distribution, the Cholesky decomposition of $(D\cdot K \times D\cdot K)$ covariance matrix, required for computing the KL divergence, might be prohibitive if $D$ and/or $K$ are large. While along the latent dimensions $d=1,\dots,D$, the prior distribution $p({\bf c}_1,\dots,{\bf c}_K)$ is factorized as in Eq.~\ref{eq:pv_olvae}. Therefore for a dimension-wise factorized encoder model, $q^c_{s,t}({\bf c}_k | G_k) = \prod_{d=1}^D q^c_{s,t}([{\bf c}_k]_d | G_k)$, a typical setting for VAEs, we are able to reduce the complexity from $\mathcal{O}((D \cdot K)^3)$ to $\mathcal{O}(D \cdot K^3)$ \cite{macneille1937partially,fattore2016partially,kim2020ordinal}. The formal last term in Eq.~\ref{eq:elbo} can be: 
\begin{equation}
\sum_{d=1}^D \textrm{KL}\Bigg( 
  \prod_{k=1}^K q^c_{s,t}([{\bf c}_k]_d | G_k) \bigg\Vert p_d([{\bf c}_1]_d,\dots,[{\bf c}_K]_d)
\Bigg).
\label{eq:content_kl}  
\end{equation}
We denote $\prod_{k=1}^K q^c_{s,t}([{\bf c}_k]_d | G_k)$ as $\mathcal{N}({\bf m}_d, {\bf S}_d)$, with ${\bf S}_d$ diagonal by definition. Each summand in Eq.~\ref{eq:content_kl} is a KL divergence between Gaussians and can be written as: 
\begin{align}
\frac{1}{2} \Bigg( 
  \textrm{Tr}({\bf C}_d^{-1} {\bf S}_d) + ({\bf a}_d - {\bf m}_d)^\top {\bf C}_d^{-1} ({\bf a}_d - {\bf m}_d) 
+ \log \frac{|{\bf C}_d|}{|{\bf S}_d|}
\Bigg), 
\label{eq:content_kl_each} 
\end{align}
The computation of inverse and determinant of ${\bf C}_d$ are tractable\footnote{The \texttt{inverse()} and \texttt{cholesky()} functions in PyTorch is used, which also allow auto-differentiations.} as $K$ is usually not a large value\footnote{E.g., $K=5$ for DR labeling system.}.


\subsection{Training processing}

The pseudo-labels $\{\hat{y}^n\}_{n=1}^{N_t}$ in target domain are iteratively selected from the reliable ordinal level predictions \cite{zou2019confidence}. Then the model is refined using the pseudo-labeled target images. Specifically, we sample $k$-class samples from both source domain ($y=k$) and target domain ($\hat{y}=k$) as a group $G_k$. We apply the conventional cross-entropy classification loss $\mathcal{L}^{CE}_{s,t}$ or the well-performed Wasserstein loss $\mathcal{L}^{W}_{s,t}$ \cite{liu2019unimodal} for the ordinal classifier prediction and the ground truth label or pseudo-label. It usually plays the most important role since we are focusing on the discriminative ordinal classification. We explicitly enforce the latent features to approaching a prior distribution as VAEs. Specifically, we minimize the KL divergence $\mathcal{L}^{KL}_{s,t}$ between $\{{\bf c}^n\}_{n=1}^{Ns},\{{\bf c}^n\}_{n=1}^{Nt}$ and RCG, and $\mathcal{L}^{KL}_{s}$ or $\mathcal{L}^{KL}_{s}$ between $\{{\bf u}_k^n\}_{n=1}^{Ns}$ or $\{{\bf u}_k^n\}_{n=1}^{Nt}$ with Gaussian. The cross domain cycle reconstruction is utilized to enforce the disentanglement of class-related/unrelated factors, and the consistent of ${\bf c}_k$ across domains. The L1 loss ($\mathcal{L}^{1}_{s}$ or $\mathcal{L}^{1}_{s}$) and adversarial loss ($\mathcal{L}^{adv}_{s}=\mathbb{E}[{\rm log}Dis_s({\bf x}_s^n) + {\rm log}Dis_s({\bf \tilde{x}}_s^n)]$ or $\mathcal{L}^{adv}_{t}=\mathbb{E}[{\rm log}Dis_{t}({\bf x}_t^n) + {\rm log}Dis_{t}({\bf \tilde{x}}_t^n)]$) are minimized to enforce the generated images to be similar as the real images in source or target domain, respectively \cite{zou2020joint,bousmalis2016domain}.

The training processing is detailed in Algorithm \ref{alg:A}. Overall, our disentangled adaptation with self-training encourages the shared encoder $Enc^c_{s,t}$ to extract discriminative and domain-invariant representations, which can generalize and facilitate ordinal classification in the target domain.

\section{Experiments}
We evaluate on real-world Diabetic Retinopathy (DR) and congenital heart disease (CHD) diagnosis tasks and demonstrate its generalizability for facial age estimation tasks. We use the convolutional layers and the first fully-connected layer of the backbone network as our feature extractors, and the remaining fully-connected layers are adopted as the classifier. The decoder has the mirror structure as encoders. We implement our methods on a V100 GPU using the PyTorch and set $\alpha,\lambda,\theta=1$, $\beta,\gamma=0.5$ for both tasks. In our experiments, the performance is not sensitive to these hyperparameters for a relatively large range. 

We use $+\mathcal{L}^W_{s,t}$ to denote the solutions using the Wasserstein loss \cite{liu2019unimodal} rather than the cross-entropy loss, while $-\mathcal{L}^{adv}_{s,t}$ indicates the ablation of the adversarial loss. 2$\sigma$ or 3$\sigma$ indicates the "two-sigma rule" or "three-sigma rule" version of RCG, respectively. For the supervised setting, we use pre-train the model with the labeled source domain, and fine-tune the model with labeled target samples as the LzyUCNN \cite{porwal2020idrid}.

\begin{table}[]
\centering
\resizebox{1\linewidth}{!}{%
\begin{tabular}{l|cccccc|c}
\hline

Method & Accuracy (\%) $\uparrow$ & QWK $\uparrow$ & MAE $\downarrow$ \\ \hline

Source only&  {48.6$\pm$0.03}  &  {45.8$\pm$0.02}& 0.61$\pm$0.01\\\hline\hline

CRST \cite{zou2019confidence}&  {56.5$\pm$0.01}  &  {53.2$\pm$0.01}& 0.50$\pm$0.02 \\
TPN \cite{pan2019transferrable}&  {56.8$\pm$0.02}  &  {52.7$\pm$0.01}& 0.52$\pm$0.01\\ 
DMRL \cite{wu2020dual}&  {57.3$\pm$0.01}  &  {53.0$\pm$0.03}& 0.50$\pm$0.02\\ 
DG-Net++ \cite{zou2020joint}&  {57.0$\pm$0.02}  &  {53.6$\pm$0.02}& 0.51$\pm$0.03\\
\hline
CRST \cite{zou2019confidence}+$\mathcal{L}^W_{s,t}$ \cite{liu2019unimodal}& 58.8$\pm$0.03  & 54.2$\pm$0.01 & 0.47$\pm$0.02 \\
DG-Net++ \cite{zou2020joint}+$\mathcal{L}^W_{s,t}$& 59.3$\pm$0.02  &  54.5$\pm$0.02& 0.48$\pm$0.02 \\\hline\hline

RCGUDA:2$\sigma$&  {61.8$\pm$0.01}  &  {56.4$\pm$0.01}& 0.41$\pm$0.02  \\ 
RCGUDA:3$\sigma$&  {61.6$\pm$0.03}  &  {56.2$\pm$0.02}& 0.43$\pm$0.01 \\

RCGUDA:2$\sigma$-$\mathcal{L}^{adv}_{s,t}$& 61.5$\pm$0.02  &  56.1$\pm$0.02& 0.42$\pm$0.02 \\
RCGUDA:2$\sigma$+$\mathcal{L}^W_{s,t}$& 62.3$\pm$0.02  &  56.9$\pm$0.01& 0.40$\pm$0.01 \\\hline\hline

Supervised \cite{porwal2020idrid}& 63.1$\pm$0.02  & 58.7$\pm$0.02& 0.38$\pm$0.01 \\\hline
\end{tabular}%
}
\caption{Experimental results for DR UDA task. $\uparrow$ larger is better.}
 
\label{tabel:dr}
\end{table}

\subsection{Diabetic Retinopathy Diagnosis}

Two public available ordinal DR datasets are adopted for DNN implementations. The Kaggle Diabetic Retinopathy (KDR)\footnote{\url{https://www.kaggle.com/c/diabetic-retinopathy-detection}} is utilized as our source domain, and the recent Indian Diabetic Retinopathy Image Dataset (IDRiD) dataset \cite{porwal2018indian} is used as our target domain. In both datasets, the diabetic retinal images are grouped following the International Clinical Diabetic Retinopathy Scale, where level 0 to 4 representing the No DR, Mild DR, Moderate DR, Severe DR, and Proliferative DR, respectively. The macular edema severity are labeled according to the occurrences of hard exudates near to macula center region. 

The KDR consists of 88,702 left or right fundus (i.e., interior surface at the back of the eye) images from 17,563 patients with the size varies from 433$\times$289 to 3888$\times$2592. Although its large scale, there is no standard field-of-view (FOV) and camera type. The images in IDRiD have the unified size (4288$\times$2848), FOV (50), and use the same Kowa VX-10$\alpha$ camera. More importantly, the IDRiD is the first dataset of Indians which has a significant shift between KDR w.r.t. population.

In the standard setting of IDRiD\footnote{\url{https://ieee-dataport.org/open-access/indian-diabetic-retinopathy-image-dataset-idrid}}, there are 413 images for training and 103 images for testing. We set its 413 training images as our $unlabeled$ target domain, while the 103 testing images are only used for UDA testing. We preprocess and resized the image to $896\times 896$ as \cite{porwal2020idrid}. The ResNet \cite{he2016deep} model in LzyUNCC \cite{porwal2020idrid} has been adopted as our backbone. Of note, LzyUNCC \cite{porwal2020idrid} uses both $labeled$ KDR and $labeled$ IDRiD for $supervised$ training, which can be regard as the ``upper bound".

\begin{figure}
  \centering
  \includegraphics[width=8.5cm]{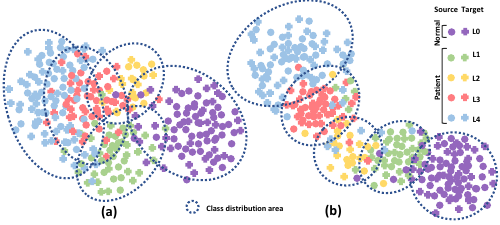}
  \caption{t-SNE visualization of the features on CHD UDA task extracted by $Enc^c_{s,t}$. Best viewed in color.}\label{fig5}  
\end{figure}

The results of the DR task are shown in Tab.~\textcolor{red}{1}. Other than the accuracy and Mean Absolute Error (MAE), we also report the quadratic weighted kappa (QWK)\footnote{\url{https://www.kaggle.com/c/diabetic-retinopathy-detection/overview/evaluation}}. QWK can reflect the misclassification proportional to the distance between the prediction and ground-truth \cite{liu2019unimodal,liu2020unimodal}. Several recent UDA methods are compared, e.g., self-training based CRST \cite{zou2019confidence}, Prototypical Network (TPN) \cite{pan2019transferrable}, disentangle-based adversarial and self-training method (DG-Net++) \cite{zou2020joint}, dual mixup regularized learning (DMRL) \cite{wu2020dual}. Although these methods have demonstrated their effectiveness in categorical classification based UDA, they do not take the inter-class correlation into account. A possible solution is applying \cite{zou2019confidence,zou2020joint} on top of the Wasserstein loss \cite{liu2019unimodal}. We note that \cite{pan2019transferrable} itself is the improvement of the loss function, which is not directly compatible with the Wasserstein loss \cite{liu2019unimodal}.

Our proposed RCGUDA outperforms these solutions consistently. Our improvements w.r.t. QWK and MAE are even more appealing. We note that QWK and MAE can be the more reasonable measure for ordinal data \cite{liu2018ordinal}, since it considers the severity of different misclassification. In Fig.~\textcolor{red}{4}, we illustrated that RCGUDA could provide a more concentrated averaged prediction distribution for the target testing samples. The 2$\sigma$ version can be slightly better than 3$\sigma$ version, considering the label noise. We also provide the ablation study of adversarial loss, which not only improves the performance but also speeds up the convergence of about 1.2 times. Furthermore, our framework is also orthogonal to the advanced ordinal classification loss, e.g., \cite{liu2019unimodal}. With the help of Wasserstein loss \cite{liu2019unimodal}, the UDA performance can approach the previously supervised LzyNUCC \cite{porwal2020idrid}.

\subsection{Real-world CHD Transfer}

Congenital heart disease (CHD) is the most prevalent congenital disability and the primary cause of death in newborns. The clinical diagnosis is usually based on the selected keyframes from the five-view echocardiogram videos. We collect a multi-center dataset of five-view echocardiograms of 1,600 labeled source subjects (L0-L4 levels) from Beijing Children's Hospital and 800 unlabeled target subjects from the Harvard Medical School. To quantify the severity levels, each subject is confirmed by either at least two senior ultrasound doctors or final intraoperative diagnosis. Note that there are discrepancies w.r.t. imaging devices (PHILIPS iE 33 vs. EPIQ 7C), patient populations, and the echocardiogram imaging experience of clinical doctors. 

We adopt the standard multi-channel network for five-view echocardiogram processing. We use 80\% and 20\% target subjects for training and testing, respectively. For comparison, we re-implement the recent state-of-the-art methods with the same backbone and experiment setting. The results are shown in Tab.~\ref{tabel:chd}. The improvements over the previous state-of-the-art are consistent with the DR UDA task. The 3$\sigma$ performs better than 2$\sigma$, since the relatively accurate label.

In Fig.~\textcolor{red}{5} left, we can see that the CHD samples tend to be distributed unordered. With our RCG prior, both the samples are ordinally grouped into the high-density region w.r.t. severity level in Fig.~\textcolor{red}{5} right. The ordinal prior can be effectively induced into the latent space features.


\begin{table}[]
\centering
\resizebox{1\linewidth}{!}{%
\begin{tabular}{l|cccccc|c}
\hline

Method & Accuracy (\%) $\uparrow$ & QWK $\uparrow$ & MAE $\downarrow$ \\ \hline

Source only& 66.2$\pm$0.02  & 60.3$\pm$0.01 & 0.42$\pm$0.02\\\hline\hline

CRST \cite{zou2019confidence}+$\mathcal{L}^W_{s,t}$ \cite{liu2019unimodal}& 72.5$\pm$0.01  & 69.9$\pm$0.02 & 0.36$\pm$0.02 \\
DG-Net++ \cite{zou2020joint}+$\mathcal{L}^W_{s,t}$& 72.4$\pm$0.01  &  70.2$\pm$0.01& 0.35$\pm$0.02 \\\hline\hline

RCGUDA:2$\sigma$&  {74.6$\pm$0.02}  &  {71.8$\pm$0.01}& 0.33$\pm$0.01  \\ 
RCGUDA:3$\sigma$&  {75.0$\pm$0.03}  &  {72.1$\pm$0.02}& 0.32$\pm$0.03 \\

RCGUDA:3$\sigma$-$\mathcal{L}^{adv}_{s,t}$& 74.5$\pm$0.01  &  71.9$\pm$0.03& 0.33$\pm$0.01 \\
RCGUDA:3$\sigma$+$\mathcal{L}^W_{s,t}$& 75.6$\pm$0.03  &  72.5$\pm$0.02& 0.31$\pm$0.01 \\\hline\hline

Supervised+$\mathcal{L}^W_{s,t}$ \cite{liu2019unimodal}& 78.4$\pm$0.02  & 74.6$\pm$0.01& 0.28$\pm$0.02 \\\hline
\end{tabular}%
}
\caption{Experimental results for CHD UDA task. $\uparrow$ larger is better.}
 
\label{tabel:chd}
\end{table}

\subsection{Face age estimation}

Although our framework has great potentials for medical image analysis, it can be generalized to the facial age estimation, which also has an ordinal label. We use MORPH Album II \cite{ricanek2006morph} as our labeled source domain, and choose IMDB-WIKI \cite{Rothe-ICCVW-2015,Rothe-IJCV-2016} as unlabeled source domain.

MORPH Album II \cite{ricanek2006morph} is a widely used dataset for estimating face age. It collects 55,134 RGB images from 13,617 people. The age is labeled between 16 to 77. Besides White or Black, the proportion of the other ethnicity is relatively low (4\%). {IMDB-WIKI dataset}\footnote{\url{https://data.vision.ee.ethz.ch/cvl/rrothe/imdb-wiki/}} 
is consisted of the celebrity faces from the IMDB and Wikipedia. Each image is labeled as the difference between the published data and the birthday of the account. In summary, the age is ranged from $5$ to $90$ years old. There are both gray and RGB images, and stamp portraits, which lead to the age label can be quite noisy and biased. Following the evaluation protocol, the face detector is applied to crop the face region for subsequent processing, and the multi-face images are dropped. Then, we have 224,418 images out of the original 523,051 images. We divide the IMDB-WIKI dataset following the subject ID-independent manner and use 50\% for the unlabeled target domain training data and 50\% for testing.

\begin{table}[]
\centering
\resizebox{1\linewidth}{!}{%
\begin{tabular}{l|cccccc|c}
\hline

Method & Accuracy (\%) $\uparrow$ & QWK $\uparrow$ & MAE $\downarrow$ \\ \hline

Source only& 48.7$\pm$0.01  & 42.8$\pm$0.02 & 0.46$\pm$0.01 \\\hline\hline

CRST \cite{zou2019confidence}+$\mathcal{L}^W_{s,t}$ \cite{liu2019unimodal}& 61.3$\pm$0.02  & 60.1$\pm$0.02 & 0.39$\pm$0.02 \\
DG-Net++ \cite{zou2020joint}+$\mathcal{L}^W_{s,t}$& 62.4$\pm$0.02  &  60.7$\pm$0.01& 0.39$\pm$0.01 \\\hline\hline

RCGUDA:3$\sigma$&  {67.8$\pm$0.02}  &  {63.2$\pm$0.02}& 0.36$\pm$0.01  \\ 
RCGUDA:2$\sigma$&  {68.1$\pm$0.02}  &  {64.5$\pm$0.03}& 0.35$\pm$0.02 \\

RCGUDA:2$\sigma$-$\mathcal{L}^{adv}_{s,t}$& 67.7$\pm$0.03  &  63.0$\pm$0.03& 0.36$\pm$0.01 \\
RCGUDA:2$\sigma$+$\mathcal{L}^W_{s,t}$& 68.3$\pm$0.02  &  64.8$\pm$0.01& 0.34$\pm$0.02 \\\hline\hline

Supervised+$\mathcal{L}^W_{s,t}$ \cite{liu2019unimodal}& 72.6$\pm$0.03  & 70.4$\pm$0.02& 0.32$\pm$0.01 \\\hline
\end{tabular}%
}
\caption{Experimental results for age UDA task. $\uparrow$ larger is better.}
 
\label{tabel:age}
\end{table}

We configure the age as the expected ordinal content factor. To unify the age groups in two datasets, we partition age values into 6 age-range classes. A age-range class involves 10 years interval,i.e., 16-25, 26-35, 36-45, 46-55, 56-65, 66-75. We note that the variations other than age, e.g., pose, illumination and facial expression, are regarded as ordinal-class unrelated (unlabeled). Following \cite{pan2018mean,liu2020unimodal}, we choose the VGG16 as our backbone. The batch size was set to 64 for VGG16. In addition, the initial learning rate is 0.001. We reduce the learning rate by multiplying 0.1 per 15 epochs. 

The results are shown in Tab.~\textcolor{red}{3}. The improvement of our RCGUDA is consistent with the medical tasks. We note that the age group in IMDB-WIKI is significantly noisier than the medical data, and demonstrates large inner-class variations. Therefore, the "two-sigma rule" version of RCG can perform better than its "three-sigma rule" version. Actually, we can also choose the decimal factor to sigma, while the performance can be stable for a large range and have the same QWK from 1.8$\sigma$ to 2.4$\sigma$ version of RCG.


\section{Conclusions}

This paper targets to explicitly induce the ordinal prior for ordinal classification UDA. We propose to meet the ordinal constraint by defining a partially ordered set (poset). Specifically, we developed a recursively conditional Gaussian (RCG) for ordered constraint modeling, which admits a tractable joint distribution prior. By enforcing RCG as the prior distribution of our joint disentanglement and adaptation framework rather than the conventional Gaussian, the performance on DR and CHD diagnosis and facial age estimation tasks can be significantly improved. The three/two-sigma setting of RCG can flexibly fit for the different degrees of label noise without the sophisticated ordinal noisy modeling. It is also orthogonal to the recently developed ordinal classification loss, which can be simply added to our RCGUDA. Our experiments evidenced that our RCG can be a powerful and versatile unsupervised solution for ordinal UDA. For future work, we plan to explore the task with a more sophisticated ordinal label, e.g., pathology level segmentation.

\section*{Acknowledgments}
The funding support from National Institute of Health (NIH), USA (NS061841, NS095986), Fanhan Tech., Jiangsu NSF (BK20200238), PolyU Central Research Grant (G-YBJW) and Hong Kong Government General Research Fund GRF (152202/14E) are greatly appreciated.


{\small
\bibliographystyle{ieee_fullname}
\bibliography{main}
}

\end{document}